%% file: main.tex
\PassOptionsToPackage{dvipsnames}{xcolor}
\documentclass{article} %
\usepackage{iclr2025_conference,times}
\usepackage{booktabs}
\usepackage{multirow}
\usepackage{xspace}
\usepackage{enumitem}
\usepackage{wrapfig}
\usepackage[colorlinks]{hyperref}

\input{math_commands.tex}

\input{packs}

\newcommand{\llava}{LLaVA\xspace}
\newcommand{\ours}{FocalLens\xspace}
\newcommand{\oursc}{FocalLens-CLIP\xspace}

\hypersetup{
citecolor=NavyBlue
}

\usepackage[textwidth=2.5cm]{todonotes}

\title{FocalLens: Instruction Tuning Enables\\Zero-Shot Conditional Image Representations}

\author{Cheng-Yu Hsieh$^{1,2}$, \
Pavan Kumar Anasosalu Vasu$^{2}$, \
Fartash Faghri$^{2}$, \
Raviteja Vemulapalli$^{2}$, \\
\bf
Chun-Liang Li$^{2}$, \
Ranjay Krishna$^{1,2}$, \
Oncel Tuzel$^{2}$, \
Hadi Pouransari$^{2}$
\\
$^{1}$University of Washington, $^{2}$Apple
}

\iclrfinalcopy %
\begin{document}

\maketitle

\begin{abstract}
\input{sections/0-abstract}
\end{abstract}

\input{sections/introduction}

\input{sections/related}

\input{sections/method_new}

\input{sections/experiments}

\input{sections/conclusion}

\newpage
\bibliography{main}
\bibliographystyle{iclr2025_conference}

\newpage
\input{sections/appendix}

\end{document}

%% file: math_commands.tex
\usepackage{amsmath,amsfonts,bm}

\def\eqref#1{equation~\ref{#1}}

\def\1{\bm{1}}

\def\vx{{\bm{x}}}
\def\vy{{\bm{y}}}

\DeclareMathAlphabet{\mathsfit}{\encodingdefault}{\sfdefault}{m}{sl}
\SetMathAlphabet{\mathsfit}{bold}{\encodingdefault}{\sfdefault}{bx}{n}

%% file: packs.tex
\usepackage{hyperref}
\usepackage{url}
\usepackage{subcaption}
\usepackage{graphicx}
\usepackage[capitalise]{cleveref}
\usepackage{color, colortbl}

%% file: sections/0-abstract.tex
Visual understanding is inherently contextual---what we focus on in an image depends on the task at hand. For instance, given an image of a person holding a bouquet of flowers, we may focus on either the person such as their clothing, or the type of flowers, depending on the context of interest. Yet, most existing image encoding paradigms represent an image as a fixed, generic feature vector, overlooking the potential needs of prioritizing varying visual information for different downstream use cases. In this work, we introduce \textit{\ours}, a \textit{conditional} visual encoding method that produces different representations for the same image based on the context of interest, expressed flexibly through natural language. We leverage vision instruction tuning data and contrastively finetune a pretrained vision encoder to take natural language instructions as additional inputs for producing conditional image representations. Extensive experiments validate that conditional image representation from \ours better pronounce the visual features of interest compared to generic features produced by standard vision encoders like CLIP. In addition, we show \ours further leads to performance improvements on a range of downstream tasks including image-image retrieval, image classification, and image-text retrieval, with an average gain of 5 and 10 points on the challenging SugarCrepe and MMVP-VLM benchmarks, respectively.

%% file: sections/introduction.tex
\section{Introduction}
In recent years, vision foundation models that are pretrained with large-scale datasets~\citep{dosovitskiy2020image, chen2022pali, radford2021learning, schuhmann2022laion} have become the cornerstone for visual feature extraction, powering downstream applications ranging from classification~\citep{dosovitskiy2020image}, segmentation~\citep{caron2021emerging}, retrieval~\citep{radford2021learning}, to multimodal large language models (MLLMs)~\citep{ramesh2021zero,li2022blip,liu2024visual,reid2024gemini, mckinzie2024mm1, driess2023palm}.
Despite the variety of pretraining schemes~\citep{radford2021learning,caron2021emerging,he2022masked,oquab2023dinov2,el2024scalable}, most commonly used vision foundation models, such as CLIP~\citep{radford2021learning}, are designed to encode the rich information contained in (a patch of) an image into a single feature vector, wherein this \textit{general} feature representation is expected to encapsulate all information that may be leveraged by various potential downstream tasks.

However, by aiming to extract \textit{general-purpose} features that can serve as many downstream tasks as possible, image representations obtained from these task-agnostic vision foundation models may inevitably compromise relevant information that is \textit{specific} to the downstream task of interest.
For instance, CLIP models are known to produce image representations that capture the high-level semantics well~\citep{radford2021learning, ramesh2021zero}, but often struggle with understanding the finer-grained details and intrinsics of the image, such as attribute associations, spatial relationships, camera perspective, and so on~\citep{vaze2023genecis, hsieh2024sugarcrepe, tong2024eyes}.

\begin{figure}[t]
    \centering
    \includegraphics[width=1\linewidth]{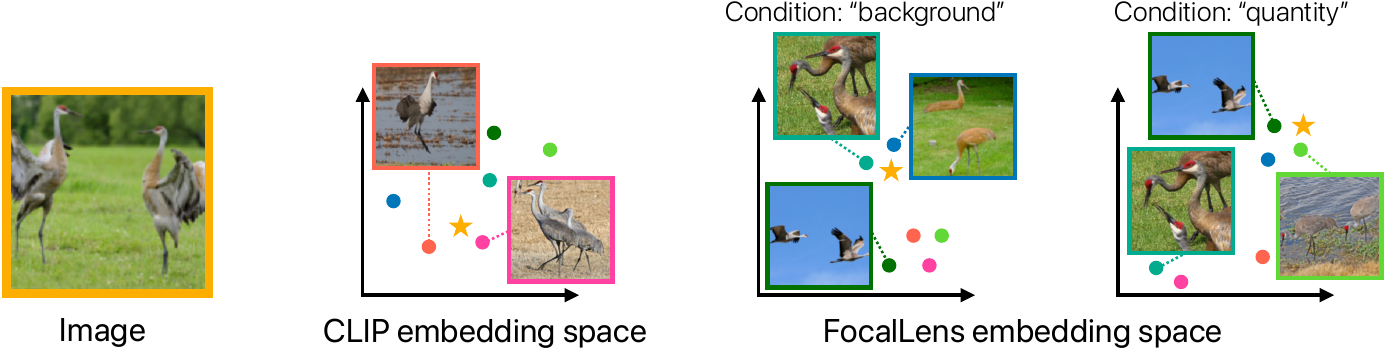}
    \caption{For a given image, the CLIP embedding space is static and structured based on overall semantics. However, \ours dynamically rearranges the embedding space based on the specified condition, bringing instances that are more similar under that condition closer together. We show the top-2 nearest neighbors for both CLIP and \ours embeddings (once conditioned on ``background'' and once on ``quantity'').}
    \label{fig:teaser}
\end{figure}

In this work, instead of aiming to learn a model that produces a fixed image representation in fulfilling different goals, we consider learning an \textit{adaptive} vision foundation model that encodes an image differently conditioned on the downstream task of interest, allowing the resultant image representations to prioritize information relevant to the specified condition over other available semantics.
Furthermore, as opposed to pre-defining the downstream tasks in a priori~\citep{clip-model-zoo-experts, wu2021fashion}, our goal is an \textit{adaptive generalist} model that is able to adapt to broad potential use cases in a \textit{zero-shot} fashion.
Specifically, we consider utilizing free-form natural language texts as a rich and flexible interface to condition~\footnote{We use ``condition (conditional)'' and ``adapt (adaptive)'' interchangeably in this paper.} the model given different downstream purposes, inspired by recent literature~\citep{wei2021finetuned, su2022one, liu2024visual}. For instance, given a task of retrieving images of similar background scene to a given query image, by specifying through the text condition: ``\textit{What is the background in the image?}'', we expect to guide the model in focusing more on the background features of the image, as illustrated in~\cref{fig:teaser}.

We introduce \emph{\ours}, a contrastive finetuning framework that transforms a pretrained vision-language model (VLM) into a text-conditioned vision encoder that is able to produce visual representations with better ``focus'' on the information relevant to the given instructions.
Specifically, leveraging visual instruction tuning dataset~\citep{dai2023instructblip,liu2024visual}, in the format of (\texttt{instruction}, \texttt{image}, \texttt{output}), \ours\ aligns the visual representation of \texttt{image} to better adhere to \texttt{instruction}, using the corresponding \texttt{output} to guide the alignment.
To demonstrate this approach, we apply \ours\ to representative pretrained MLLM and vision encoder: \llava~\citep{liu2024visual} and CLIP~\citep{radford2021learning}, and name the resultant text-conditioned vision encoder models \emph{\ours-MLLM} and \emph{\oursc} respectively, as illustrated in \cref{fig:setups}.

Through extensive evaluations on over 60 tasks, we observe that \ours models demonstrate a strong ability to condition representations based on the given text instructions, significantly outperforming existing baselines like CLIP. On average, \ours achieves up to 9 points higher performance, with even greater improvements on specific tasks, for image-image retrieval tasks.
In addition, when used in downstream applications, \ours's conditional image representations further lead to clear gains compared to existing baselines. For instance, on image-text retrieval benchmarks, we show an average improvements of 5 and 10 points respectively on SugarCrepe~\citep{hsieh2024sugarcrepe} and MMVP-VLM~\citep{tong2024eyes}, comparing favorably to other CLIP models that are much larger (up to 2.5$\times$) in size.
On image classification, \ours also shows superior performances than CLIP, especially in low-data regime.
Finally, further qualitative study showcases various intriguing application scenarios that can be supported by \ours.

%% file: sections/related.tex
\section{Related work}
\paragraph{Foundation models for vision encoding.}
Modern vision foundation models trained on web-scale datasets~\citep{dosovitskiy2020image,jia2021scaling,schuhmann2022laion,oquab2023dinov2} are used as the common underlying visual feature extractor to produce image representations that drive various downstream applications~\citep{radford2021learning,ramesh2021zero,kirillov2023segment,zhou2022extract}.
While there are many pretraining objectives~\citep{oquab2023dinov2,he2022masked,el2024scalable,radford2021learning}, existing schemes typically train the vision models to produce a single ``general'' image representation that hopefully captures all relevant information contained in the given image, or utilize information derived from diverse captions to help learning more discriminative image features~\citep{lavoie2024modeling}.
Nonetheless, as an image naturally contains rich and dense information, a fixed and general-purpose representation may not sufficiently pronounce information relevant to specific downstream contexts of interest~\citep{kar2024brave,wang2024sam,tong2024eyes,hsieh2024sugarcrepe}.
Our work aims to learn vision encoder that is capable of extracting different representations from a single image conditioned on downstream use cases at test-time, different from universal image embedding approaches that aim to learn a universal model for different domains without explicit conditioning~\citep{google2023universalembedding, ypsilantis2023towards}.

\paragraph{Conditional vision representations.}
Implicit and task-specific conditioning of visual features have been studied in the literature~\citep{liu2024visual,dai2023instructblip,tong2024cambrian,eftekhar2023selective,vani2024sparo,team2024chameleon}.
For instance, the hidden representations in MLLMs may be interpreted as a type of conditional image representation, where the visual features are fused with text instructions for producing different output responses.
Nonetheless, conditional visual representations considered in prior work are designed specifically to their model and respective applications, \emph{e.g.}, generative conversations~\citep{dai2023instructblip} and embodied AI~\citep{eftekhar2023selective}.
In this work, we are interested in conditional visual representations that may be used for various downstream applications, such as classifications, image-image or image-text retrieval.

\paragraph{Vision-language joint representation learning.}
There is a rich literature in vision-language (joint) representation learning~\citep{lu2019vilbert, li2019universal, kim2021vilt, radford2021learning, jiang2024e5}. Our work is related as we aim for a model that can comprehend both images and natural language conditions.
Concurrent to ours, recent works~\citep{jiang2024e5, jiang2024vlm2vec} consider MLLM's output space as a universal representation space for both vision and language inputs. Nonetheless, in addition to the MLLM-based approach, we study an alternative promising CLIP-based approach with comprehensive analysis which leads to various performance benefits.
Relatedly, composed image retrieval~\citep{wu2021fashion, saito2023pic2word, zhang2024magiclens} considers developing models of underlying similar capabilities that generate image embeddings given both image and text. However, different from our goal to use text conditioning to extract downstream-specific \textit{intrinsic} visual features, their goal is to \textit{extrinsically} ``compose'' semantics from both texts and images, largely towards image-retrieval purposes.

%% file: sections/method_new.tex
\section{Conditional embeddings via instruction contrastive tuning}
\label{sec:method}
Our goal is to develop an adaptive vision foundation model that is capable of encoding an image into tailored embeddings conditioned on the downstream task of interest, as specified through natural language texts.

We consider the visual instruction tuning data~\citep{liu2024visual}, which covers diverse tasks, and has demonstrated great generalization of MLLMs in different benchmarks. The visual instruction tuning data is in the triplet format of (\texttt{image}, \texttt{instruction}, \texttt{output}). For instance, given an image of ``a Yorkshire Terrier wearing a green cloth'',  the  \texttt{output} is ``The dog is wearing a green cloth with strawberry prints on it'' with the \texttt{instruction} ``What is the dog wearing?''. Alternatively, when the instruction  is ``What is the type of the dog'', the output is ``The dog is a Yorkshire Terrier'' correspondingly. 
MLLMs~\citep{dai2023instructblip,liu2024visual} leverage the triplet instruction tuning for text generation: given (\texttt{image}, \texttt{instruction}), generating \texttt{output}. Instead, we propose to utilize contrastive learning~\citep{radford2021learning} on the triplet instruction tuning data. Specifically, given an image encoder conditioned on the \textit{instruction}, we match the output embedding with a text embedding of \textit{output}. We call the proposed method as \emph{FocalLens}, which leverage instruction tuning data to \emph{contrastively} tune the pretrained image encoder, such that it can better focus on desired information and generalize to diverse downstream tasks. We explore tuning two different representative vision-language models with \ours: MLLMs (\cref{sec:mllm_model}) and CLIP (\cref{sec:clip_model}).

\begin{figure}[t]
	\centering
	\begin{subfigure}[t]{0.45\linewidth}
		\centering
		\includegraphics[width=\linewidth]{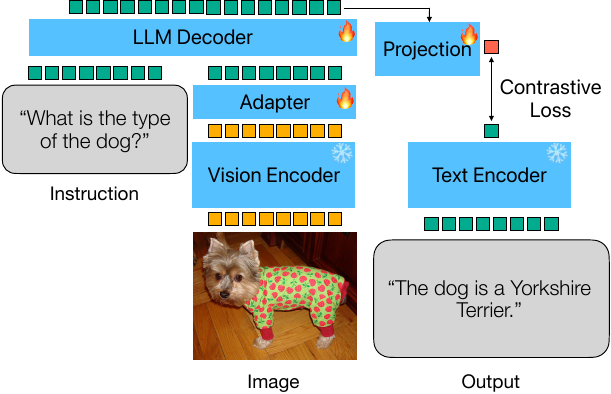}
		\caption{\ours-MLLM}
		\label{fig:llava-setup}
	\end{subfigure}
	\hspace{4mm}
	\begin{subfigure}[t]{0.45\linewidth}
		\centering
		\includegraphics[width=\linewidth]{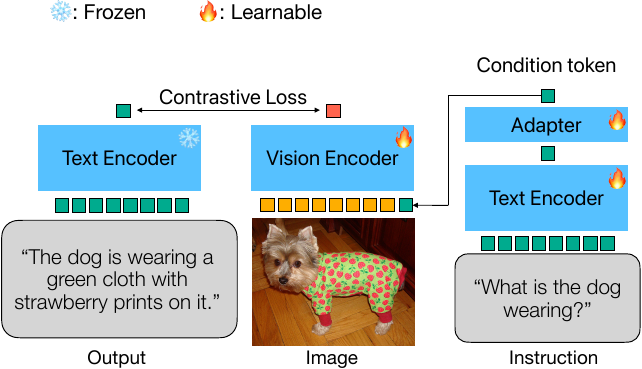}
		\caption{\ours-CLIP}
		\label{fig:clip-setup}
	\end{subfigure}
	\caption{\ours is applied to two vision-language models to extract text-conditioned visual features: {\bf (a)} modifying Llava-like VLMs, which already have text-conditioning capabilities, to produce a global visual feature, and {\bf (b)} modifying ViT~\citep{dosovitskiy2020image} based CLIP-like VLMs, which already produce a global visual feature, to condition their output feature based on a text condition.}
	\label{fig:setups}
\end{figure}

\subsection{\ours with MLLMs}
\label{sec:mllm_model}
MLLMs~\citep{liu2024visual, dai2023instructblip} generate textual responses regarding an image based on the given input text instructions. Given (\texttt{instruction}, \texttt{image}), the goal is to generate \texttt{output}.
However, as the original model objective is text generation rather than producing explicit representation for downstream tasks, the conditional visual information may be dispersed throughout the model, and there is no direct access to them by design.

In \ours,  instead of training the MLLM to generate \texttt{output} given (\texttt{image}, \texttt{instruction}) as in the original auto-regressive objective, we append a special indicator token \texttt{<eos\_token>} to MLLM's input sequence, and consequently train the indicator's output token to align with the CLIP text embedding of the targeted output in a constrative manner.  Here, we use an off-the-shelf frozen CLIP text encoder to obtain the target output embedding. With the contrastive objective, we encourage the model to condense information relevant to the image, with the specified instructions, into a single output representation.
We show the overall model architecture in~\cref{fig:llava-setup}.

\subsection{\ours with CLIP}
\label{sec:clip_model}

Unlike MLLMs, CLIP models by design generate image representations~\citep{radford2021learning}, where these image embeddings are already widely utilized in a variety of downstream tasks~\citep{ramesh2021zero, liu2024visual}.
However, CLIP models are inherently limited to producing a fixed representation for each image, regardless of the downstream task of interest.
Although strong in capturing high-level semantics, these general visual features are shown to lack various aspects of fine-grained image details that can be critical for downstream tasks~\citep{hsieh2024sugarcrepe, tong2024eyes}.
To tackle this, we propose to make CLIP's vision encoder task-aware, such that it is able to adapt its representations based on specific requirements, thereby capturing specific aspects of the image essential for different applications.

To incorporate natural language instructions into CLIP's vision encoder, we consider first converting \texttt{instruction} into a ``condition text embedding'', which is then treated as an additional token that is fed into the image encoder alongside the standard image tokens and the CLS token. Afterwards, the model is trained as in standard CLIP using a contrastive loss, aligning the resultant text-conditioned image representations with their corresponding textual outputs.
By instruction tuning, we aim to allow the vision encoder to generalize to a broad range of scenarios of interest that can be described via natural language at test-time~\citep{wei2021finetuned, su2022one}.
We illustrate the \ours-CLIP training setup in~\cref{fig:clip-setup}.

%% file: sections/experiments.tex
\section{Experiments}
In this section, we first demonstrate the benefits of conditional image representations (\cref{sec:exp_exploratory}) over the generic representations produced by CLIP, using a toy dataset.
We then extensively evaluate \ours models' capability in characterizing downstream conditions on a variety of tasks, compared to existing baselines (\cref{sec:benchmarks}).
By zooming in on \ours-CLIP, we demonstrate that its conditional image representations improve performance across a range of downstream tasks, including image-text retrieval, image classification, and image-image retrieval (\cref{sec:downstream_eval}).

\paragraph{Setup.} We train \ours models with the visual instruction tuning data used in \llava~\citep{liu2024visual}. The dataset contains around 150k examples, wherein 60k examples are multi-turn conversations and thus can be treated as multiple triplets of (\texttt{image}, \texttt{instruction}, \texttt{output}), where the image remains the same. During training, we expand conversation data within batches to encourage models to output different representations given the same image but different instructions.
For \ours-MLLM, we follow the training recipe of \llava~\citep{liu2024visual} to obtain a base MLLM before further training with the proposed contrastive loss. For \oursc, we initialize the base CLIP model with OpenAI's CLIP-ViT-L-14-336~\citep{radford2021learning}, which is also the underlying vision encoder used in \llava. We initialize the additional text encoder for instructions to have the same weight as the original text encoder.

For contrastive instruction tuning, given a batch of triplet instruction data $(\vx^{(i)}_{\text{img}}, \vx^{(i)}_{\text{ins}}, \vy^{(i)})$, where $\vy^{(i)}$ is the expected output for sample $i$, we form the pair-wise similarity matrix $S$, such that
\begin{equation}
S_{i,j} = \phi(\vx^{(i)}_{\text{img}}, \vx^{(i)}_{\text{ins}} ) ^T \mathcal{T} (\vy^{(j)}),
\end{equation}
where $\phi$ is the encoding process that produce the conditional image embedding from both image $\vx_{\text{img}}$ and instruction $\vx_{\text{ins}}$, 
and $\mathcal{T}$ is the (frozen) text encoder that generates the target embedding from $\vy$.
We apply scaled Softmax to the rows of similarity matrix and compute the contrastive loss following CLIP~\citep{radford2021learning}.
We report further training details in \cref{app:setup}. In addition, we report all prompts used for conditioning \ours models during evaluation in \cref{app:prompts}.

\paragraph{Image-image retrieval as an evaluation protocol.}
We consider the common image-image retrieval evaluation to measure the quality of image representations produced from different vision encoders~\citep{google2023universalembedding, caron2021emerging}.
Specifically, given a query image, image-image retrieval tasks the model to retrieve other images from a gallery that are ``similar'' to the query image.
We are especially interested in the scenario wherein the very definition of ``similar'' changes as the downstream tasks vary~\citep{vaze2023genecis}.
To facilitate such evaluations, we adopt datasets where we may define various similarities between images based on \textit{test-time} interest determined through a text condition. We introduce these datasets in the following sections.
For each dataset, when not otherwise specified, we report mean Average Precision (mAP) as the evaluation metric.

\subsection{Conditional  representations better characterize task-specific details}
\label{sec:exp_exploratory}
We empirically validate the benefits of having the flexibility to encode an image based on the given condition of interest over using a fixed representation when downstream purpose varies, as considered in most prevailing vision encoding paradigms~\citep{radford2021learning, caron2021emerging}.
Here, we restrict ourselves to a toy dataset to demonstrate the idea, and we shall expand our studies in the following sections.

\begin{figure}[t!]
\begin{minipage}{0.35\textwidth}
        \centering
        \includegraphics[width=\linewidth]{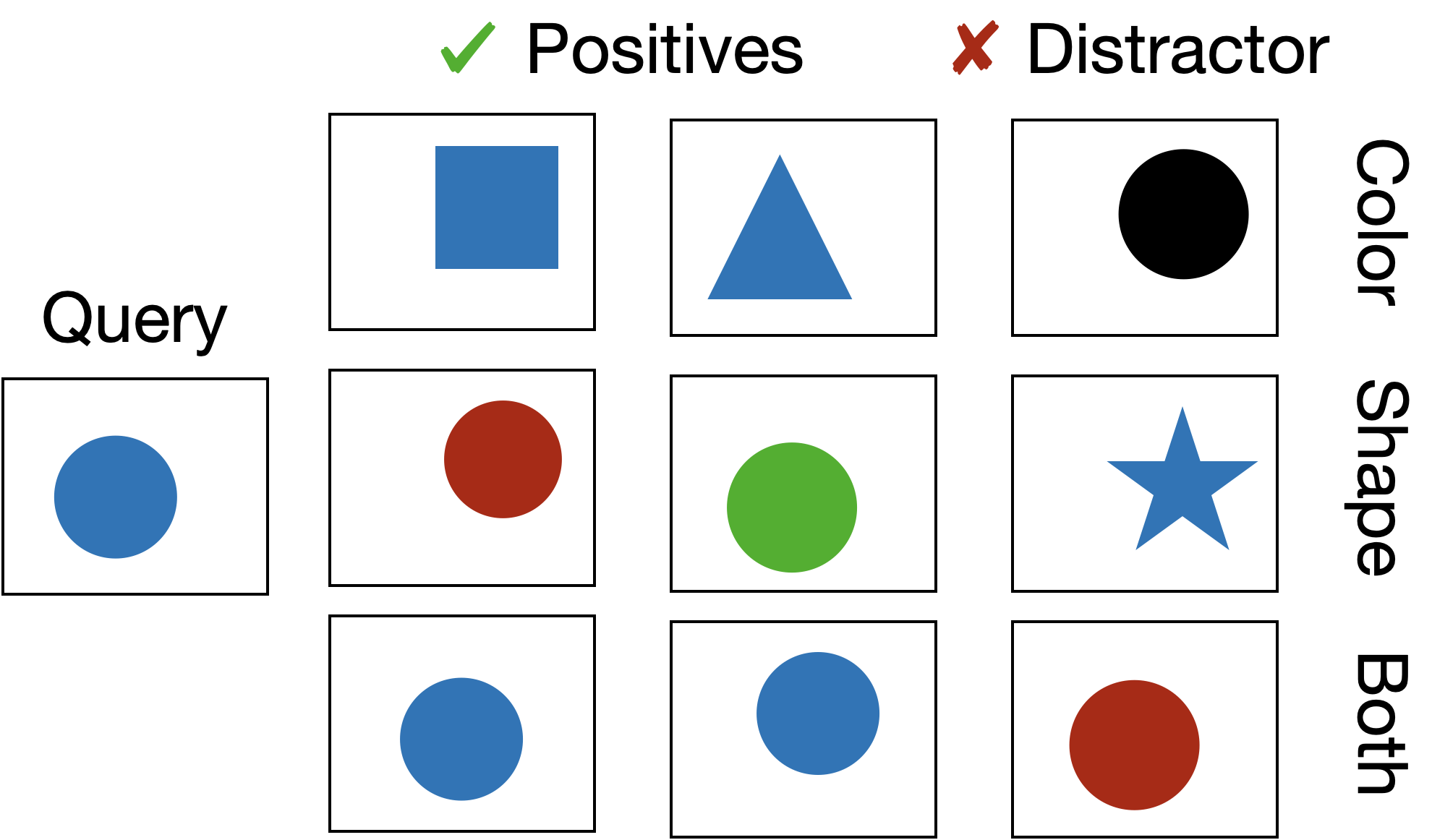}
        \caption{ColorShape examples with a query image, three conditions, and corresponding positives and distractors.}
        \label{fig:colorshape_example}
\end{minipage}
\hspace{0.025\textwidth}
\begin{minipage}{0.625\textwidth}
  \centering
  \small
\resizebox{\textwidth}{!}{
\begin{tabular}{l ccccc} 
    \toprule
     & \multicolumn{4}{c}{ColorShape} & \multirow{2}{0.8cm}{\centering Cont. Color} \\
     \cmidrule(lr){2-5}   
     Model & {Color} & {Shape} & {Both} & Avg. & \\
    
    \midrule
    
    CLIP (task-agnostic) & 57.10 & \underline{90.24} & \underline{99.36} & \cellcolor{gray!25} 82.23 & 0.158 \\
    \midrule
    \ours-MLLM & {\bf 99.94} & 82.56 & 98.92 & \cellcolor{gray!25} {\bf 93.80} & {\bf 0.560}\\
    \ours-CLIP & \underline{87.28} & {\bf 93.51} & {\bf 99.99} & \cellcolor{gray!25} \underline{93.59} & \underline{0.405}\\
\bottomrule 
\end{tabular}
}
\captionsetup{type=table}
\caption{Image-image retrieval results on ColorShape dataset. Conditional representations from \ours better capture the given conditions compared to the task-agnostic representations of CLIP.}
\label{tab:colorshape}
\end{minipage}
\end{figure}

{\bf A toy ColorShape dataset.} ColorShape is a synthetic dataset where each image contains a certain colored shape. There are in total 4 different colors and shapes respectively. We generate 500 different images with random position and size of the object for each combination of color and shape. At test-time, we may define the intent for retrieval based on different aspects. Specifically, we may group each image into different categories based on either only its color, only its shape, or both. \cref{fig:colorshape_example} shows some examples from the dataset.

The pretrained CLIP model~\citep{radford2021learning} serves as the standard encoder baseline where the image representations are fixed even when the test-time condition varies. For the conditional vision encoders, we consider both \ours-MLLM and \ours-CLIP models discussed in \cref{sec:method}.
We show their retrieval performances on the ColorShape dataset when the test-time condition varies.

{\bf Non-adapative image representations overlook specific aspects of images.}
From \cref{tab:colorshape}, on the simple ColorShape dataset, CLIP yields almost perfect retrieval performances when we define image categories based on both color and shape.
However, in the context where we are specifically interested in categorizing images based only on the color, CLIP's performance drops significantly to 57 mAP point. On the other hand, when we define similarity based only on shape, CLIP achieves relatively better performances at 90 mAP point. Combining the results, while CLIP can produce \textit{general} representation that is strong at grouping objects of certain shape and color together, its overall representation space is biased towards the ``shape'' of objects, and much less discriminative over the ``color'' aspect.
This also echos the observations made in recent works~\citep{tong2024eyes,hsieh2024sugarcrepe}, suggesting that CLIP's representation, while powerful for general tasks, may overlook fine-grained details such as color, highlighting a need for approaches to better adapt and capture the nuanced visual characteristics, depending on the task at hand.

{\bf Conditional image representations better capture information relevant to the downstream task.}
In \cref{tab:colorshape}, as opposed to CLIP model, the conditional image representations produced from  both adaptive vision encoders, the MLLM-based and the CLIP-based model, achieve much more balanced (and superior) results than CLIP's representation when the downstream condition varies. When averaged across three different scenarios (``color'', ``shape'', and ``both''), both conditional vision encoders improve over 10 mAP point compared to CLIP.
The conditional CLIP-based model also always outperforms CLIP, when evaluated separately on the three respective conditions.

In addition to using discrete color labels (e.g., “red”, “blue”) to define image similarity, we also consider a more sophisticated setup where image similarity is measured based on L2 distance in RGB space. Specifically, in this Continuous Color variant, we assign randomly sampled RGB colors to the objects. During evaluation, our goal is to retrieve images with colors closer to that of the query image. We compute the rank correlation between the similarity measured in the model’s image representation space and the ground-truth similarity defined in RGB space. In this setup, both \ours models significantly outperform CLIP as show in the last column of \cref{tab:colorshape}.

\subsection{\ours improves image representations across benchmarks}
\label{sec:benchmarks}
Using the ColorShape toy dataset, we validated the benefits of adapting image representations for downstream tasks. We now compare \ours to existing vision encoders and relevant baselines across a comprehensive set of evaluation benchmarks.

\paragraph{Evaluation benchmarks.} We consider a total of 49 different tasks across 4 coarse-grained categories in our evaluation suite as briefly described below. We include dataset details in \cref{app:dataset}.

\begin{itemize}[left=0pt]
    \item \textbf{CelebA-Attribute}~\citep{liu2015faceattributes}: CelebA is a dataset consisting of celebrity face images. Each face image is associated with various properties spanning from the hair color of the person, the eyebrow shape, to whether the person is wearing eyeglasses, and so on. We vary the downstream condition of interest across different properties for retrieval. For instance, when conditioned on ``eyeglasses'' with a query image showing a person is (not) wearing eyeglasses, the model is tasked to retrieve other face images with (without) eyeglasses.
    We manually select a total of 29 different properties that can be objectively labeled, and exclude more subjective properties such as ``attractiveness'' or ``young''.
    We notice that the class within each attribute may be imbalanced, resulting in high mAP even with random guess. We thus report scaled performances  w.r.t. random guess by: $\frac{p-r}{1-r}$, where $p$ is the original mAP and $r$ is the random guess mAP.
    
    \item \textbf{GeneCIS}~\citep{vaze2023genecis}: GeneCIS presents various image retrieval tasks for evaluating conditional image similarity. Given a query image (``a white laptop'') and a condition (``color''), the goal is to retrieve the most similar image (another ``white laptop'') from a gallery that contains implicitly similar distractors with wrong conditions (e.g., ``a black laptop''). We report the ``Focus attribute'' and ``Focus object'' tasks from GeneCIS. As each query image contains only a single positive in the gallery, we report Recall@3 following prior work~\citep{zhang2024magiclens}. 

    \item \textbf{ImageNet-Subset}~\citep{Deng2009ImageNetAL}: In addition to the above benchmarks with specific downstream conditions of interest, we as well evaluate our models on standard ImageNet classes, where the condition corresponds to the image ``classes'' as defined by ImageNet.
    Specifically, we create 14 different retrieval sub-tasks based on coarse-grained categories from WordNet~\citep{wordnet} hierarchy (e.g., ball, bird, dog, etc.). In each task (e.g., dog), the goal is to retrieve images (from all dog images) with the same type of instance (same breed of dog) as the query image.
    
    \item \textbf{Fine-grained classification datasets}: Similar to ImageNet, we incorporate 4 finer-grained classification datasets, including Oxford Flowers~\citep{nilsback2008automated}, Stanford Cars~\citep{krause20133d}, FGVC Aircraft~\citep{maji2013fine}, and Food-101~\citep{bossard2014food}.
    
\end{itemize}

\paragraph{Baselines.} We consider CLIP~\citep{radford2021learning} as the task-agnostic vision encoder model. We also compare to models that are able to generate conditional visual representations, including the Q-former used in InstructBLIP~\citep{li2023blip, dai2023instructblip}, and MagicLens~\citep{zhang2024magiclens} that is designed specifically for composed image-retrieval with open-ended instructions. We include details of the baselines in \cref{app:baseline}.

\begin{table}[t]
  \centering
  \small
  \caption{Results on CelebA-Attribute and GeneCIS.}
\resizebox{\textwidth}{!}{
\begin{tabular}{l cccccccc} 
    \toprule 
     & \multicolumn{5}{c}{CelebA-Attribute} & \multicolumn{3}{c}{GeneCIS} \\
     \cmidrule(lr){2-6} \cmidrule(lr){7-9}
     Model & Blond Hair & Smiling & Wavy Hair & Lipstick & Avg. 29 tasks & Attribute & Object & Avg. \\
    
    \midrule

    CLIP & 6.20 & 8.68 & 7.54 & 41.45 & \cellcolor{gray!25} 13.59 & 43.10 & 25.81 & \cellcolor{gray!25} 34.46 \\
InstructBLIP & 21.03 & 21.71 & 13.91 & 34.64 & \cellcolor{gray!25} 16.19 & {\bf 47.00} & 34.03 & \cellcolor{gray!25} \underline{40.52} \\
MagicLens & 8.24 & 9.98 & 10.76 & 54.12 & \cellcolor{gray!25} 13.42 & 39.00 & \underline{35.50} & \cellcolor{gray!25} 37.25\\
\midrule
\ours-MLLM & \underline{25.76} & {\bf 34.43} & {\bf 17.61} & {\bf 68.07} & \cellcolor{gray!25} {\bf 22.67} & \underline{45.35} & 30.20 & \cellcolor{gray!25} 37.78\\
\ours-CLIP & {\bf 32.22} & \underline{22.11} & \underline{16.89} & \underline{62.50} & \cellcolor{gray!25} \underline{21.32} & {\bf 43.30} & {\bf 43.72} &\cellcolor{gray!25} {\bf 43.51}\\

\bottomrule 
\end{tabular}
}
\label{tab:celeba_genecis}
\end{table}

\begin{table}[t]
  \centering
  \small
  \caption{Results on ImageNet-Subset and fine-grained classification datasets.}
\resizebox{\textwidth}{!}{
\begin{tabular}{l cccccccccc} 
    \toprule 
     & \multicolumn{5}{c}{ImageNet-Subset} & \multicolumn{5}{c}{Fine-grained classification datasets} \\
     \cmidrule(lr){2-6} \cmidrule(lr){7-11}
     Model & Ball & Cat & Dog & Fish & Avg. 14 tasks & Flower & Car & Aircraft & Food & Avg.\\
    
    \midrule

    CLIP & 64.63 & 53.00 & 16.55 & \underline{61.79} & \cellcolor{gray!25} 51.03 & {\bf 83.87} & \underline{45.14} & {\bf 25.96} & 58.66 & \cellcolor{gray!25} \underline{53.41} \\
InstructBLIP & 66.44 & 51.22 & 9.60 & 59.16 & \cellcolor{gray!25} 47.67 & \underline{80.26} & 25.97 & 13.47 & 54.32 & \cellcolor{gray!25} 43.51 \\
MagicLens & 68.10 & 50.14 & 17.28 & 58.84 & \cellcolor{gray!25} 46.36 & 74.88 & 23.95 & 17.55 & {\bf 65.13} & \cellcolor{gray!25} 45.38 \\
\midrule
\ours-MLLM & {\bf 78.99} & \underline{53.24} & \underline{29.25} & 57.40 & \cellcolor{gray!25} \underline{52.34} & 43.92 & 18.59 & 14.73 & 50.93 & \cellcolor{gray!25} 32.04 \\
\ours-CLIP & \underline{70.01} & {\bf 56.80} & {\bf 33.15} & {\bf 65.37} & \cellcolor{gray!25} {\bf 55.29} & 80.23 & {\bf 54.72} & \underline{21.44} & \underline{64.16} & \cellcolor{gray!25} {\bf 55.14} \\

\bottomrule 
\end{tabular}
}
\label{tab:classification_tasks}
\end{table}

{\bf \ours improves significantly over existing baselines given specific downstream conditions.}
From \cref{tab:celeba_genecis}, both variants of \ours provide significant gains over the task-agnostic CLIP baseline on CelebA-Attribute and GeneCIS, when there are specific conditions to respect.
We see an overall gain of 9 points on CelebA-Attribute.
Looking more closely at the individual conditions on CelebA-Attribute (complete results reported in \cref{app:results}), we observe that when the condition of interest is  ``smiling'', we see a significant gap of 26 points between CLIP and \ours, where the gap is as large as 48 points on certain attributes.
Similarly on the GeneCIS benchmark, by specifying the attribute such as color or certain object to focus on, \ours improves over CLIP by an average of 9 points.

On CelebA-Attribute and GeneCIS, we also see \ours models demonstrate outperforming (or favorable) results when compared to prior task-aware vision encoders (i.e., InstructBLIP and MagicLens), that are also given the downstream condition of interest when generating the image representations. Specifically, \ours-CLIP achieves the best overall performances, winning over the stronger InstructBLIP baseline by 5 and 3 points respectively on CelebA-Attribute and GeneCIS, validating the effectiveness of our proposed strategy.

{\bf \ours maintains or improves over existing baseline on generic conditions.}
In Table~\ref{tab:classification_tasks}, we compare model performances on ImageNet-Subset and the fine-grained classification datasets, where the downstream goal is generic instance classification.
First, CLIP model demonstrates competitive performances on both ImageNet-Subset and fine-grained classification tasks, showing that its embeddings are indeed strong at representing generic features when it comes to standard ``type'' classification.
In contrast, InstructBLIP and MagicLens suffer performance drops on both ImageNet-Subset and fine-grained tasks.
On the other hand, we see \ours (especially \ours-CLIP) maintains comparable performances to CLIP on fine-grained datasets and attains even better performances on ImageNet-Subset. We explain the improvement on ImageNet by that conditioning \ours with instructions such as ``What is the type of dog?'' helps the model to better focus on the specific object of interest but not other potential distractors in the image (e.g., the ``toy'' besides the dog).

\subsection{Comparative analysis of \ours variants}
\label{sec:llava_vs_clip}
Both \ours-MLLM and \ours-CLIP yield promising results in the experiments.
One major difference between \ours-MLLM and \ours-CLIP is their underlying pretrained models' output modality.
Specifically, the original MLLM model in \ours-MLLM is trained to autoregressively produce textual outputs, while CLIP's vision encoder is trained to produce image embeddings.
We are thus interested in understanding whether this difference affects the underlying characteristics of the output representations in \ours-MLLM and \ours-CLIP.

\begin{wraptable}{r}{0.4\textwidth}
  \centering
  \small
  \caption{Comparison between \ours-MLLM and \ours-CLIP on fuzzy conditions with CelebA-Identity.}
  \begin{tabular}{l c} 
    \toprule
    Model & CelebA-Identity \\
    \midrule
    \ours-MLLM & 14.48 \\
    \ours-CLIP & {\bf 46.84} \\
    \bottomrule
  \end{tabular}
  \label{tab:fuzzy_condition}
\end{wraptable}

To test this, we consider downstream conditions that require visual features beyond semantic concepts that are describable by text.
In particular, on CelebA, instead of considering conditions such as whether the person is wearing glasses or not, which is answerable in simple words (``yes'' or ``no''), we consider a \textit{fuzzy} condition where the image similarity is defined by the \textit{identity} of the person. Textual representations that do not carry visual information may fail at achieving good performance on this task, as identity is hardly describable through natural language. 

In \cref{tab:fuzzy_condition}, we observe that \ours-MLLM suffers from a clear performance gap compared to \ours-CLIP. This suggests that \ours-MLLM may rely more on MLLM's original textual output modality, which is limited for tasks requiring rich visual information. Similar observations are also hinted by its relatively low performance on fine-grained classification results in \cref{tab:classification_tasks}.
In contrast, \ours-CLIP, with its underlying model being a vision encoder, is better suited for tasks requiring richer visual detail.
Based on this observation, we focus on \ours-CLIP for the remainder of the experiments.

\subsection{\ours representations improve downstream applications}
\label{sec:downstream_eval}
In addition to evaluations based only on image representations, we show how image representations produced from \ours-CLIP can drive improvement on downstream tasks including image-text retrieval and image classification in a low-data regime where only a small amount of downstream task data is available for training.

\begin{table}[t]
  \centering
  \small
  \begin{minipage}{\textwidth}
  \caption{Image-Text Retrieval on SugarCrepe for vision-language compositionality evaluation.}
  \vspace{-1mm}
      \resizebox{\textwidth}{!}{
\begin{tabular}{l ccccccccc} 
    \toprule
    & \multicolumn{8}{c}{SugarCrepe} \\
    \cmidrule(lr){2-9}
    Model & Replace-obj & Replace-att & Replace-rel & Swap-obj & Swap-att & Add-obj & Add-att & Avg.\\

    \midrule
    
    OpenAI ViT-L-14~(\citeyear{radford2021learning}) & 94.49 & 80.58 & 66.78 & 64.08 & 62.46 & 80.74 & 74.27 &\cellcolor{gray!25} 74.77 \\
    OpenAI RN50x64~(\citeyear{radford2021learning}) & 94.49 & 83.50 & 70.63 & 61.79 & \underline{66.67} & 83.27 & 73.99 &\cellcolor{gray!25} 76.33 \\
    LAION ViT-g-14~(\citeyear{schuhmann2022laion}) & {\bf 95.76} & {\bf 85.03} & \underline{72.40} & \underline{63.01} & {\bf 71.17} & {\bf 91.51} & \underline{82.08} &\cellcolor{gray!25} {\bf 80.14} \\
    
    \midrule 
    
    \ours-CLIP & \underline{95.64} & \underline{84.51} & {\bf 75.53} & {\bf 65.30} & 66.36 & \underline{86.12} & {\bf 83.09} & \cellcolor{gray!25} \underline{79.51}\\

\bottomrule 
\end{tabular}
}
\label{tab:sugarcrepe}
  \end{minipage}

  \vspace{2mm}
  
  \begin{minipage}{\textwidth}
    \caption{Image-Text Retrieval on MMVP-VLM.}
    \vspace{-1mm}
\resizebox{\textwidth}{!}{
\begin{tabular}{l cccccccccc} 
    \toprule
  
    & \multicolumn{10}{c}{MMVP-VLM} \\
    \cmidrule(lr){2-11}
    Model & Orientation & Presence & State & Quantity & Spatial & Color & Structure & Text & Camera & Avg.\\
    \midrule

    OpenAI ViT-L-14~(\citeyear{radford2021learning}) & 6.7 & 20.0 & 26.7 & 6.7 & \underline{13.3} & 33.3 & {\bf 46.7} & \underline{20.0} & 13.3 &\cellcolor{gray!25} 20.7 \\

    MetaCLIP ViT-H-14~(\citeyear{xu2023demystifying}) & 6.7 & 13.3 & {\bf 60.0} & \underline{13.3} & 6.7 & \underline{53.3} & \underline{26.7} & 13.3 & {\bf 33.3} &\cellcolor{gray!25} \underline{25.2} \\
    
    EVA01 ViT-g-14~(\citeyear{sun2023eva}) & 6.7 & \underline{26.7} & \underline{40.0} & 6.7 & \underline{13.3} & {\bf 66.7} & 13.3 & 13.3 & \underline{20.0} &\cellcolor{gray!25} 23.0 \\

    \midrule
    \ours-CLIP & 6.7 & {\bf 33.3} & 33.3 & {\bf 40.00} & {\bf 26.7} & {\bf 66.7} & 20.0 & {\bf 26.7} & \underline{20.0} & \cellcolor{gray!25} {\bf 30.4} \\

\bottomrule 
\end{tabular}
}
\label{tab:mmvp}
  \end{minipage}

\end{table}

{\bf Image-text retrieval.}
A prevailing usage of image representations is to enable cross-modality retrieval. Here, we include two image-text prediction benchmarks, where the goal is to predict the correct textual description of a given image.
Specifically, we adopt SugarCrepe~\citep{hsieh2024sugarcrepe} and MMVP-VLM~\citep{tong2024eyes}.
SugarCrepe presents challenging hard-negative text distrators along with a positive description for the model to select from, where existing models are shown to struggle with.
Similarly, MMVP-VLM particularly collects examples with visual patterns where CLIP vision encoder are shown to fall short.

In \cref{tab:sugarcrepe} on SugarCrepe, we compare \oursc to several standard CLIP models of different sizes, and trained with different data sizes.
First, compared to the underlying CLIP model used in \oursc (i.e., OpenAI ViT-L-14), \oursc achieves around 4.7 point improvements on average, with consistent improvements across all different sub-tasks with individual gains up to 9 points on Replace-rel and Add-att. Interestingly, the two sub-tasks test the model's capability in understanding fine-grained relationships and attributes in the image, where standard CLIP models struggle the most~\citep{hsieh2024sugarcrepe}. This suggests \oursc's image representations are able to better characterize fine-grained visual details. Furthermore, by scaling up the model size from 428M to 623M, the RN50x64 model still underperform our smaller \oursc model (551M for both image and text encoders). On the other hand, \oursc shows competitive performances compared to the 2.5$\times$ bigger ViT-g-14 model trained on 5$\times$ more data.

From \cref{tab:mmvp} on MMVP-VLM, we see \oursc significantly outperforms the baseline ViT-L-14 model consistently across all sub-tasks, by an average of 9.7 points. Furthermore, we note that our \oursc model also compares favorably to the much larger ViT-H-14 (1.8$\times$ larger) and ViT-g-14 (2$\times$ larger) on individual sub-tasks, where \oursc achieves the best overall performance with a lead of 5.2 point.

\begin{wrapfigure}{r}{0.4\textwidth}
  \centering
  \includegraphics[width=\linewidth]{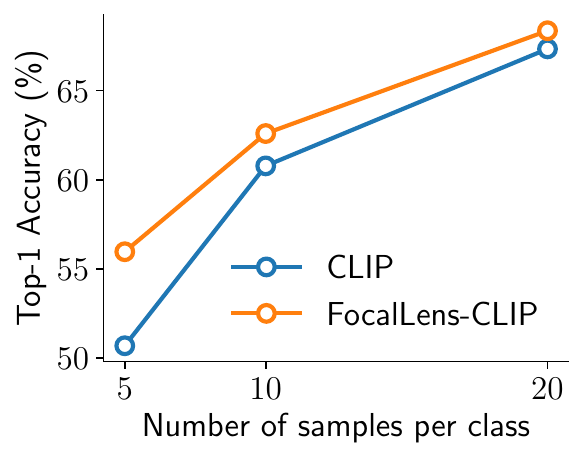}
  \caption{Linear probing results comparing CLIP and \oursc.}
  \label{fig:probing}
\end{wrapfigure}

\paragraph{Linear probing in low-data regime.}
We evaluate the performance of \oursc in a linear probing setup, where only a small amount of downstream task data is available for training. We use the largest dataset in ImageNet-Subset introduced in \cref{sec:benchmarks}, focusing on different dog breeds (a total of 118 classes). In the low-data setup~\citep{henaff2020data,luo2017label,vemulapalliknowledge}, we assume there are $k$ instances available for each class for training and consider $k=5, 10, 15$. We freeze the backbone and replace the CLIP projection layer with a linear layer to perform 118-way classification. The linear probe is trained for 100 epochs following prior works like~\citep{dBOT}. We sweep over learning rates from 1e-2 to 1e-4 in steps of 2.5e-3 and report the performance of the best checkpoint. We compare \oursc to OpenAI ViT-L-14 in this setup, as shown in \cref{fig:probing}. In the extreme setting, where only 5 instances per class is available to train the linear probe, \oursc outperforms CLIP-ViT-L by 5.3\%. This result further reinforces our observation that conditional image representations are more efficient in extracting information relevant to downstream tasks.

\begin{figure}[!t]
    \centering
    \includegraphics[width=\linewidth]{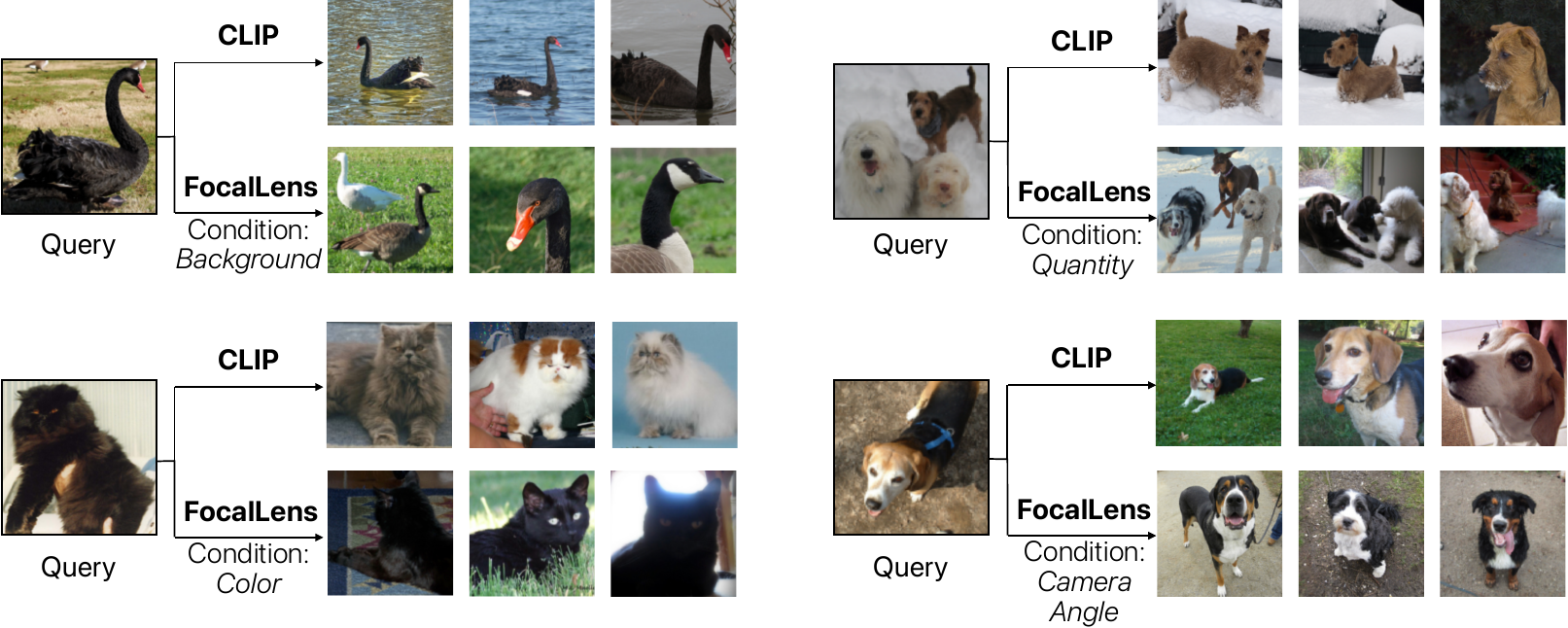}
    \caption{Comparison between CLIP and \oursc on conditional image retrieval.}
    \label{fig:qualitative}
\end{figure}

{\bf Qualitative analysis on conditional image-retrieval.}
We qualitatively compare the top-$k$ images retrieved by using \oursc's conditional image embeddings with those retrieved by standard CLIP, specifically when given various downstream conditions.
For this qualitative study, we treat all images in the 14 coarse-grained categories considered in ImageNet-Subset as the gallery for retrieval.
In \cref{fig:qualitative}, we showcase several intriguing examples across various aspects of conditioning \oursc captures.
In the top-left example, we consider a scenario where we are interested in retrieving images of similar background to the query image. Given the query image of ``a goose on a grassy field'', although the images retrieved by CLIP do all contain goose, all images have the background of water instead of grassy field. Conversely, we see images retrieved by \oursc all have similar grassy background as expected.
Similarly, in the top-right, we see \oursc faithfully reflects the interested condition of quantity, retrieving images with 3 dogs as in the query image, whereas images retrieved by CLIP is largely based on their instance type (same species of dog), and cannot reflect the downstream interest. More examples demonstrate that color or even implicit visual features such as camera angle can also be characterized by \oursc.

%% file: sections/conclusion.tex
\section{Conclusion}

In this work, we introduced \ours, a zero-shot conditional visual embedding model that focuses the representation on specific aspects of the image described in the given text. \ours is trained using existing visual instruction tuning datasets to align the conditional image representation with the textual description. Experiments on a comprehensive set of tasks, including image-to-image retrieval, image classification, and image-to-text retrieval, demonstrate that \ours matches or exceeds the performance of state-of-the-art models. 

\paragraph{Limitations.}

Although experiments demonstrate that \ours can be effectively trained using visual instruction tuning datasets, model performance could be enhanced by designing customized datasets for this task, which we leave for future study. Moreover, the relatively small scale of the visual instruction tuning datasets may hinder alignment accuracy for highly specialized concepts that are entirely absent from the dataset.

%% file: sections/appendix.tex
\appendix
\section{Datasets}
\label{app:dataset}

\paragraph{CelebA-Attribute.} There are a total of 40 different binary attributes in CelebA dataset~\citep{liu2015faceattributes}, from which we select 29 attributes we consider objective, including: ``Arched Eyebrows'', ``Bags Under Eyes'', ``Bald'', ``Bangs'', ``Black Hair'', ``Blond Hair'', ``Brown Hair'', ``Gray Hair'', ``Blurry'', ``Bushy Eyebrows'', ``Double Chin'', ``Eyeglasses'', ``Goatee'', ``Male'', ``Mouth Slightly Open'', ``Mustache'', ``No Beard'', ``Oval Face'', ``Pale Skin'', ``Rosy Cheeks'', ``Sideburns'', ``Smiling'', ``Straight Hair'', ``Wavy Hair'', ``Wearing Earrings'', ``Wearing Hat'', ``Wearing Lipstick'', ``Wearing Necklace'', ``Wearing Necktie''.

\paragraph{ImageNet-Subset.}
The ImageNet dataset~\citep{Deng2009ImageNetAL} is organized according to the nouns in the WordNet hierarchy~\citep{wordnet} and consists of 1000 classes. To evaluate the performance of conditioned representations, we form multiple subsets of ImageNet using the intermediate nodes from the WordNet hierarchy. We list all the ImageNet subsets we created in \cref{tab:imagenet_subsets_details}.

\begin{table}[h]
\small
\caption{ImageNet-Subset datasets and number of classes per each.}
 \label{tab:imagenet_subsets_details}
\centering
\resizebox{0.99\columnwidth}{!}
    {
    \begin{tabular}{c|cccccccccccccc}
    \toprule
         \multirow{2}{*}{\centering \bf Node Name}&\multirow{2}{*}{Dog}&\multirow{2}{*}{Bird}&\multirow{2}{1.4cm}{\centering Musical Instrument}&\multirow{2}{*}{Snake}&\multirow{2}{*}{Fish}&\multirow{2}{*}{Monkey}&\multirow{2}{*}{Ball}&\multirow{2}{*}{Car}&\multirow{2}{1.1cm}{\centering Edible Fruit}&\multirow{2}{*}{Beetle} & \multirow{2}{*}{Cat}&\multirow{2}{*}{Spider} &\multirow{2}{*}{Bag}&\multirow{2}{*}{Piano}\\
         &&&&&&&&&&&&&&\\
    \midrule
         {\bf Num classes}& 118&59&28&17&16&13&10&10&10&8&7&6&5&2\\
    \bottomrule
    \end{tabular}
    }
\end{table}

\section{Baselines}
\label{app:baseline}

\paragraph{CLIP.}
We consider CLIP as a task-agnostic vision encoder baseline. In all experiments, we use OpenAI's CLIP-ViT-L-patch14-336 released checkpoint~\citep{radford2021learning}. The model size is 428M including both vision and text encoder. We consider the same model checkpoint in \ours-MLLM and \oursc.

\paragraph{InstructBLIP.}
InstructBLIP~\citep{dai2023instructblip} is a MLLM that connects a frozen vision encoder, CLIP~\citep{fang2023eva}, to a large language model (LLM) decoder to enable multi-modal capabilities. Specifically, it adopts an instruction-aware Q-former architecture~\citep{li2023blip} as the connector. The Q-former takes in as input the image embedding extracted from the underlying vision encoder, along with tokenized text instructions. Through cross-attention design, the Q-former outputs multiple instruction-aware image tokens to be fed into the LLM decoder. In our experiments, we average over all image tokens to obtain the image representation used in our evaluations. We use the same instructions as in \ours for conditioning InstructBLIP.

\paragraph{MagicLens.}
MagicLens~\citep{zhang2024magiclens} is a model trained specifically for composed image retrieval with a web-scale 36M-sized dataset. The model takes in both a reference image and natural language text to produce image representations that composes the semantics from both the input image and text. In our experiments, we condition MagicLens model using the same text instructions used for \ours.

\section{Experiment details}
\label{app:setup}
\paragraph{Computation resource.}
We train \ours models on single node machines with 8 A100 GPUs.

\paragraph{Hyperparameters.}
For contrastive training with \ours, we report the hyperparameters used in \cref{tab:hparam}.

\begin{table}[h]
  \centering
  \small
  \caption{Training hyperparameters.}
  \begin{tabular}{l ccccc} 
    \toprule
    Model & Batch size & Epoch & Learning rate  & Weight decay & Warmup ratio \\
    \midrule
    \ours-MLLM & 384 & 2 & 2e-5 & 0. & 0.03 \\
    \oursc & 2048 & 20 & 2e-5 & 0 & 0.03 \\
    \bottomrule
  \end{tabular}
  \label{tab:hparam}
\end{table}

\section{Instructions used for different tasks}
\label{app:prompts}
Here, we detail the instructions we use for different tasks for conditioning \ours and other instruction-aware baselines.

\begin{table}[h]
    \centering
    \caption{Instructions and templates used for different datasets and conditions.}
    \resizebox{\textwidth}{!}{
    \begin{tabular}{ll l}
        \toprule
        \textbf{Dataset} & \textbf{Condition} & \textbf{Instruction} \\
        \midrule
        ColorShape & Color & What is the color of the object in the image?\\
         & Shape &  What is the shape of the object in the image? \\
         &  Both &  What is the color and shape of the object in the image? \\
        
        \midrule
        CelebA-Attribute & Noun attributes (e.g., Arched Eyebrows) & Does the person in the image have \{attribute\}? \\
         & Adjective attributes (e.g., Bald) &  Is the person in the image \{attribute\}?\\
        
        \midrule
        
         CelebA-Identity & - & Gender, age, eye color, hair color, face shape, facial hair of the person. \\

         \midrule
        GeneCIS & Focus attribute & Focus on the \{attribute\}. \\
         & Focus object & Is there \{object\}? \\

        \midrule
        ImageNet-Subset & category (e.g., dog) & What type of \{category\} is in the image?\\

         \midrule
        Fine-grained datasets & category (e.g., flower) & What type of \{category\} is in the image? \\
    
         \midrule
        SugarCrepe & Replace-obj & Focus on the presence of objects in the image. \\
         & Replace-att & Focus on the color, patterns and other attributes of the objects in the image. \\
         & Replace-rel & What are the relationships between the objects in the image? \\
         & Swap-obj & What are the actions, states, colors, patterns and relationships of the objects in the image? \\
         & Swap-att & What kind of objects are in the image? \\
         & Add-obj & What is not in the image? \\
         & Add-att & What is not in the image? \\

        \midrule
        MMVP-VLM  & Orientation & Describe the orientation, position, or the direction of the object. \\
         & Presence & Focus on the presence of objects in the image. \\
         & State & Focus on the specific state or the condition of the objects in the image. \\
         & Quantity & Focus on the quantity of the objects in the image. \\
         & Spatial & Describe the spatial relationship and the positions of the objects in the image. \\
         & Color & Focus on the color of the objects in the image. \\
         & Structural & Describe the state of the objects in the image. \\
         & Text & Focus on the texts on the objects in the image. \\
         & Camera & Describe the perspective and view from which the photo is taken. \\
         
        \bottomrule
    \end{tabular}
    }
    \label{tab:all_prompts}
\end{table}

\section{Full experiment results}
\label{app:results}

\subsection{CelebA-Attribute full results}
We report full CelebA-Attribute results in \cref{tab:celeba-all}.
\begin{table}[h]
  \centering
  \small
  \caption{Full results on CelebA-Attribute.}
\resizebox{\textwidth}{!}{
\begin{tabular}{l cccccccccc} 
    \toprule
    Model & Arched Eyebrows & Bags Under Eyes & Bald & Bangs & Black Hair & Blond Hair & Blurry & Brown Hair & Bushy Eyebrows & Double Chin \\

    \midrule

  CLIP  & 8.13 & 12.00 & 24.52 & 2.86 & 7.96 & 6.20 & 5.52 & -0.58 & 11.98 & 18.35 \\
InstructBLIP & 7.12 & 8.35  & 27.40 & 4.95 & 9.50 & 21.03 & 14.67 & -0.81  & 3.73  & 11.01 \\
MagicLens & 11.32 & 12.10 & 15.14 & 2.44 & 7.48 & 8.24 & 8.95 & -3.22 & 6.75  & 13.88 \\
\ours-MLLM & 15.15 & 14.98 & 19.23 & 4.38 & 17.95 & 25.76 & 6.14 & 4.44  & 6.88  & 15.37 \\
\oursc & 13.38 & 13.00 & 26.68 & 8.19 & 10.24 & 32.22 & 11.03 & 5.53  & 9.99  & 15.94 \\

    \midrule
    
Model &  Eyeglasses & Goatee & Gray Hair & Male & Mouth Slightly Open & Mustache & No Beard & Oval Face & Pale Skin & Rosy Cheeks \\

    \midrule
CLIP & 17.84 & 20.16 & 24.19 & 54.55 & 4.72 & 20.92 & 27.64 & 1.63 & 3.22 & -3.15 \\
InstructBLIP & 41.83 & 16.17 & 22.56 & 43.66 & 12.87 & 19.16 & 23.75 & 0.77 & 2.73 & -3.45 \\
MagicLens & 15.52 & 11.28 & 20.13 & 64.56 & 6.04 & 13.50 & 27.52 & 1.83 & 1.98 & 1.95 \\
\ours-MLLM & 47.72 & 20.96 & 22.40 & 96.82 & 33.41 & 19.30 & 34.30 & 1.66 & 1.36 & 5.85 \\
\oursc & 24.90 & 29.04 & 23.86 & 95.04 & 10.82 & 26.59 & 41.80 & 0.94 & 4.58 & -0.90 \\

    \midrule

    Model & Sideburns & Smiling & Straight Hair & Wavy Hair & Wearing Earrings & Wearing Hat & Wearing Lipstick & Wearing Necklace & Wearing Necktie \\
\midrule

 CLIP   & 18.21 & 8.68  & 3.47  & 7.54  & 7.32  & 17.60 & 41.45 & -0.67 & 21.81 \\
InstructBLIP & 12.10 & 21.71 & 3.17  & 13.91 & 13.51 & 45.11 & 34.64 & 1.94  & 36.56 \\
MagicLens & 11.54 & 9.98  & 2.84  & 10.76 & 10.92 & 21.19 & 54.12 & 3.58  & 16.97 \\
\ours-MLLM & 20.02 & 34.43 & 4.50  & 17.61 & 21.54 & 34.32 & 68.07 & 5.05  & 37.86 \\
\oursc & 32.35 & 22.11 & 2.81  & 16.89 & 12.39 & 33.58 & 62.50 & 3.07  & 29.80 \\

\bottomrule 
\end{tabular}
}
\label{tab:celeba-all}
\end{table}

\subsection{ImageNet-Subset full results}

We report full ImageNet-Subset results in \cref{tab:imagenet-all}.
\begin{table}[h]
  \centering
  \small
  \caption{Full results on ImageNet-Subset.}
\begin{tabular}{l ccccccc} 
    \toprule
    Model & Bag & Ball & Beetle & Bird & Car & Cat & Dog \\
    \midrule
CLIP & 55.61 & 64.63 & 51.84 & 66.72 & 57.73 & 53.00 & 16.55 \\
InstructBLIP & 60.13 & 66.44 & 51.10 & 45.86 & 60.54 & 51.22 & 9.60  \\
MagicLens & 53.22 & 68.10 & 43.37 & 51.69 & 54.15 & 50.14 & 17.28 \\
\ours-MLLM & 63.95 & 78.99 & 41.44 & 54.14 & 54.46 & 53.24 & 29.25 \\
\oursc & 59.44 & 70.01 & 46.88 & 64.62 & 61.84 & 56.80 & 33.15 \\

    \midrule
    Model & Fruit & Fish & Monkey & Music Instrument & Piano & Snake & Spider \\
    \midrule
CLIP & 60.95 & 61.79 & 37.79 & 39.18 & 61.97 & 32.03 & 54.61 \\
InstructBLIP & 49.74 & 59.16 & 27.96 & 41.44 & 66.17 & 26.45 & 51.61 \\
MagicLens & 57.40 & 58.84 & 26.82 & 41.18 & 57.40 & 25.74 & 43.76 \\
\ours-MLLM & 65.98 & 57.40 & 34.81 & 57.83 & 57.14 & 29.47 & 54.69 \\
\oursc & 69.78 & 65.37 & 38.30 & 61.29 & 60.60 & 32.06 & 53.93 \\

\bottomrule 
\end{tabular}
\label{tab:imagenet-all}
\end{table}

%% file: main.bbl
\begin{thebibliography}{56}
\providecommand{\natexlab}[1]{#1}
\providecommand{\url}[1]{\texttt{#1}}
\expandafter\ifx\csname urlstyle\endcsname\relax
  \providecommand{\doi}[1]{doi: #1}\else
  \providecommand{\doi}{doi: \begingroup \urlstyle{rm}\Url}\fi

\bibitem[Bossard et~al.(2014)Bossard, Guillaumin, and Van~Gool]{bossard2014food}
Lukas Bossard, Matthieu Guillaumin, and Luc Van~Gool.
\newblock Food-101--mining discriminative components with random forests.
\newblock In \emph{Computer vision--ECCV 2014: 13th European conference, zurich, Switzerland, September 6-12, 2014, proceedings, part VI 13}, pp.\  446--461. Springer, 2014.

\bibitem[Caron et~al.(2021)Caron, Touvron, Misra, J{\'e}gou, Mairal, Bojanowski, and Joulin]{caron2021emerging}
Mathilde Caron, Hugo Touvron, Ishan Misra, Herv{\'e} J{\'e}gou, Julien Mairal, Piotr Bojanowski, and Armand Joulin.
\newblock Emerging properties in self-supervised vision transformers.
\newblock In \emph{Proceedings of the IEEE/CVF international conference on computer vision}, pp.\  9650--9660, 2021.

\bibitem[{Chameleon Team}(2024)]{team2024chameleon}
{Chameleon Team}.
\newblock Chameleon: Mixed-modal early-fusion foundation models.
\newblock \emph{arXiv preprint arXiv:2405.09818}, 2024.

\bibitem[Chen et~al.(2022)Chen, Wang, Changpinyo, Piergiovanni, Padlewski, Salz, Goodman, Grycner, Mustafa, Beyer, et~al.]{chen2022pali}
Xi~Chen, Xiao Wang, Soravit Changpinyo, AJ~Piergiovanni, Piotr Padlewski, Daniel Salz, Sebastian Goodman, Adam Grycner, Basil Mustafa, Lucas Beyer, et~al.
\newblock Pali: A jointly-scaled multilingual language-image model.
\newblock \emph{arXiv preprint arXiv:2209.06794}, 2022.

\bibitem[Dai et~al.(2023)Dai, Li, Li, Tiong, Zhao, Wang, Li, Fung, and Hoi]{dai2023instructblip}
Wenliang Dai, Junnan Li, D~Li, AMH Tiong, J~Zhao, W~Wang, B~Li, P~Fung, and S~Hoi.
\newblock Instructblip: Towards general-purpose vision-language models with instruction tuning. arxiv 2023.
\newblock \emph{arXiv preprint arXiv:2305.06500}, 2, 2023.

\bibitem[Deng et~al.(2009)Deng, Dong, Socher, Li, Li, and Fei-Fei]{Deng2009ImageNetAL}
Jia Deng, Wei Dong, Richard Socher, Li-Jia Li, K.~Li, and Li~Fei-Fei.
\newblock Imagenet: A large-scale hierarchical image database.
\newblock \emph{2009 IEEE Conference on Computer Vision and Pattern Recognition}, pp.\  248--255, 2009.
\newblock URL \url{https://api.semanticscholar.org/CorpusID:57246310}.

\bibitem[Dosovitskiy(2020)]{dosovitskiy2020image}
Alexey Dosovitskiy.
\newblock An image is worth 16x16 words: Transformers for image recognition at scale.
\newblock \emph{arXiv preprint arXiv:2010.11929}, 2020.

\bibitem[Driess et~al.(2023)Driess, Xia, Sajjadi, Lynch, Chowdhery, Ichter, Wahid, Tompson, Vuong, Yu, et~al.]{driess2023palm}
Danny Driess, Fei Xia, Mehdi~SM Sajjadi, Corey Lynch, Aakanksha Chowdhery, Brian Ichter, Ayzaan Wahid, Jonathan Tompson, Quan Vuong, Tianhe Yu, et~al.
\newblock Palm-e: An embodied multimodal language model.
\newblock \emph{arXiv preprint arXiv:2303.03378}, 2023.

\bibitem[Eftekhar et~al.(2023)Eftekhar, Zeng, Duan, Farhadi, Kembhavi, and Krishna]{eftekhar2023selective}
Ainaz Eftekhar, Kuo-Hao Zeng, Jiafei Duan, Ali Farhadi, Ani Kembhavi, and Ranjay Krishna.
\newblock Selective visual representations improve convergence and generalization for embodied ai.
\newblock \emph{arXiv preprint arXiv:2311.04193}, 2023.

\bibitem[El-Nouby et~al.(2024)El-Nouby, Klein, Zhai, Bautista, Toshev, Shankar, Susskind, and Joulin]{el2024scalable}
Alaaeldin El-Nouby, Michal Klein, Shuangfei Zhai, Miguel~Angel Bautista, Alexander Toshev, Vaishaal Shankar, Joshua~M Susskind, and Armand Joulin.
\newblock Scalable pre-training of large autoregressive image models.
\newblock \emph{arXiv preprint arXiv:2401.08541}, 2024.

\bibitem[Fang et~al.(2023)Fang, Wang, Xie, Sun, Wu, Wang, Huang, Wang, and Cao]{fang2023eva}
Yuxin Fang, Wen Wang, Binhui Xie, Quan Sun, Ledell Wu, Xinggang Wang, Tiejun Huang, Xinlong Wang, and Yue Cao.
\newblock Eva: Exploring the limits of masked visual representation learning at scale.
\newblock In \emph{Proceedings of the IEEE/CVF Conference on Computer Vision and Pattern Recognition}, pp.\  19358--19369, 2023.

\bibitem[{Google Research}(2023)]{google2023universalembedding}
{Google Research}.
\newblock Introducing the google universal image embedding challenge, April 2023.
\newblock URL \url{http://research.google/blog/introducing-the-google-universal-image-embedding-challenge/}.
\newblock Accessed: 2024-09-30.

\bibitem[He et~al.(2022)He, Chen, Xie, Li, Doll{\'a}r, and Girshick]{he2022masked}
Kaiming He, Xinlei Chen, Saining Xie, Yanghao Li, Piotr Doll{\'a}r, and Ross Girshick.
\newblock Masked autoencoders are scalable vision learners.
\newblock In \emph{Proceedings of the IEEE/CVF conference on computer vision and pattern recognition}, pp.\  16000--16009, 2022.

\bibitem[Henaff(2020)]{henaff2020data}
Olivier Henaff.
\newblock Data-efficient image recognition with contrastive predictive coding.
\newblock In \emph{International conference on machine learning}, pp.\  4182--4192. PMLR, 2020.

\bibitem[Hsieh et~al.(2024)Hsieh, Zhang, Ma, Kembhavi, and Krishna]{hsieh2024sugarcrepe}
Cheng-Yu Hsieh, Jieyu Zhang, Zixian Ma, Aniruddha Kembhavi, and Ranjay Krishna.
\newblock Sugarcrepe: Fixing hackable benchmarks for vision-language compositionality.
\newblock \emph{Advances in neural information processing systems}, 36, 2024.

\bibitem[Jia et~al.(2021)Jia, Yang, Xia, Chen, Parekh, Pham, Le, Sung, Li, and Duerig]{jia2021scaling}
Chao Jia, Yinfei Yang, Ye~Xia, Yi-Ting Chen, Zarana Parekh, Hieu Pham, Quoc Le, Yun-Hsuan Sung, Zhen Li, and Tom Duerig.
\newblock Scaling up visual and vision-language representation learning with noisy text supervision.
\newblock In \emph{International conference on machine learning}, pp.\  4904--4916. PMLR, 2021.

\bibitem[Jiang et~al.(2024{\natexlab{a}})Jiang, Song, Zhang, Huang, Deng, Sun, Zhang, Wang, and Zhuang]{jiang2024e5}
Ting Jiang, Minghui Song, Zihan Zhang, Haizhen Huang, Weiwei Deng, Feng Sun, Qi~Zhang, Deqing Wang, and Fuzhen Zhuang.
\newblock E5-v: Universal embeddings with multimodal large language models.
\newblock \emph{arXiv preprint arXiv:2407.12580}, 2024{\natexlab{a}}.

\bibitem[Jiang et~al.(2024{\natexlab{b}})Jiang, Meng, Yang, Yavuz, Zhou, and Chen]{jiang2024vlm2vec}
Ziyan Jiang, Rui Meng, Xinyi Yang, Semih Yavuz, Yingbo Zhou, and Wenhu Chen.
\newblock Vlm2vec: Training vision-language models for massive multimodal embedding tasks.
\newblock \emph{arXiv preprint arXiv:2410.05160}, 2024{\natexlab{b}}.

\bibitem[Kar et~al.(2024)Kar, Tonioni, Poklukar, Kulshrestha, Zamir, and Tombari]{kar2024brave}
O{\u{g}}uzhan~Fatih Kar, Alessio Tonioni, Petra Poklukar, Achin Kulshrestha, Amir Zamir, and Federico Tombari.
\newblock Brave: Broadening the visual encoding of vision-language models.
\newblock \emph{arXiv preprint arXiv:2404.07204}, 2024.

\bibitem[Kim et~al.(2021)Kim, Son, and Kim]{kim2021vilt}
Wonjae Kim, Bokyung Son, and Ildoo Kim.
\newblock Vilt: Vision-and-language transformer without convolution or region supervision.
\newblock In \emph{International conference on machine learning}, pp.\  5583--5594. PMLR, 2021.

\bibitem[Kirillov et~al.(2023)Kirillov, Mintun, Ravi, Mao, Rolland, Gustafson, Xiao, Whitehead, Berg, Lo, et~al.]{kirillov2023segment}
Alexander Kirillov, Eric Mintun, Nikhila Ravi, Hanzi Mao, Chloe Rolland, Laura Gustafson, Tete Xiao, Spencer Whitehead, Alexander~C Berg, Wan-Yen Lo, et~al.
\newblock Segment anything.
\newblock In \emph{Proceedings of the IEEE/CVF International Conference on Computer Vision}, pp.\  4015--4026, 2023.

\bibitem[Krause et~al.(2013)Krause, Stark, Deng, and Fei-Fei]{krause20133d}
Jonathan Krause, Michael Stark, Jia Deng, and Li~Fei-Fei.
\newblock 3d object representations for fine-grained categorization.
\newblock In \emph{Proceedings of the IEEE international conference on computer vision workshops}, pp.\  554--561, 2013.

\bibitem[Lavoie et~al.(2024)Lavoie, Kirichenko, Ibrahim, Assran, Wildon, Courville, and Ballas]{lavoie2024modeling}
Samuel Lavoie, Polina Kirichenko, Mark Ibrahim, Mahmoud Assran, Andrew~Gordon Wildon, Aaron Courville, and Nicolas Ballas.
\newblock Modeling caption diversity in contrastive vision-language pretraining.
\newblock \emph{arXiv preprint arXiv:2405.00740}, 2024.

\bibitem[Li et~al.(2019)Li, Duan, Fang, Gong, Jiang, and Zhou]{li2019universal}
G~Li, N~Duan, Y~Fang, M~Unicoder-VL Gong, D~Jiang, and M~Unicoder-VL Zhou.
\newblock A universal encoder for vision and language by cross-modal pre-training. arxiv 2019.
\newblock \emph{arXiv preprint arXiv:1908.06066}, 2019.

\bibitem[Li et~al.(2022)Li, Li, Xiong, and Hoi]{li2022blip}
Junnan Li, Dongxu Li, Caiming Xiong, and Steven Hoi.
\newblock Blip: Bootstrapping language-image pre-training for unified vision-language understanding and generation.
\newblock In \emph{International conference on machine learning}, pp.\  12888--12900. PMLR, 2022.

\bibitem[Li et~al.(2023)Li, Li, Savarese, and Hoi]{li2023blip}
Junnan Li, Dongxu Li, Silvio Savarese, and Steven Hoi.
\newblock Blip-2: Bootstrapping language-image pre-training with frozen image encoders and large language models.
\newblock In \emph{International conference on machine learning}, pp.\  19730--19742. PMLR, 2023.

\bibitem[Liu et~al.(2024{\natexlab{a}})Liu, Li, Wu, and Lee]{liu2024visual}
Haotian Liu, Chunyuan Li, Qingyang Wu, and Yong~Jae Lee.
\newblock Visual instruction tuning.
\newblock \emph{Advances in neural information processing systems}, 36, 2024{\natexlab{a}}.

\bibitem[Liu et~al.(2024{\natexlab{b}})Liu, Zhou, Kong, Lin, and Ji]{dBOT}
Xingbin Liu, Jinghao Zhou, Tao Kong, Xianming Lin, and Rongrong Ji.
\newblock Exploring target representations for masked autoencoders.
\newblock In \emph{International Conference on Learning Representations}, 2024{\natexlab{b}}.

\bibitem[Liu et~al.(2015)Liu, Luo, Wang, and Tang]{liu2015faceattributes}
Ziwei Liu, Ping Luo, Xiaogang Wang, and Xiaoou Tang.
\newblock Deep learning face attributes in the wild.
\newblock In \emph{Proceedings of International Conference on Computer Vision (ICCV)}, December 2015.

\bibitem[Lu et~al.(2019)Lu, Batra, Parikh, and Lee]{lu2019vilbert}
Jiasen Lu, Dhruv Batra, Devi Parikh, and Stefan Lee.
\newblock Vilbert: Pretraining task-agnostic visiolinguistic representations for vision-and-language tasks.
\newblock \emph{Advances in neural information processing systems}, 32, 2019.

\bibitem[Luo et~al.(2017)Luo, Zou, Hoffman, and Fei-Fei]{luo2017label}
Zelun Luo, Yuliang Zou, Judy Hoffman, and Li~F Fei-Fei.
\newblock Label efficient learning of transferable representations acrosss domains and tasks.
\newblock \emph{Advances in neural information processing systems}, 30, 2017.

\bibitem[Maji et~al.(2013)Maji, Rahtu, Kannala, Blaschko, and Vedaldi]{maji2013fine}
Subhransu Maji, Esa Rahtu, Juho Kannala, Matthew Blaschko, and Andrea Vedaldi.
\newblock Fine-grained visual classification of aircraft.
\newblock \emph{arXiv preprint arXiv:1306.5151}, 2013.

\bibitem[McKinzie et~al.(2024)McKinzie, Gan, Fauconnier, Dodge, Zhang, Dufter, Shah, Du, Peng, Weers, et~al.]{mckinzie2024mm1}
Brandon McKinzie, Zhe Gan, Jean-Philippe Fauconnier, Sam Dodge, Bowen Zhang, Philipp Dufter, Dhruti Shah, Xianzhi Du, Futang Peng, Floris Weers, et~al.
\newblock Mm1: Methods, analysis \& insights from multimodal llm pre-training.
\newblock \emph{arXiv preprint arXiv:2403.09611}, 2024.

\bibitem[Miller(1995)]{wordnet}
George~A Miller.
\newblock Wordnet: a lexical database for english.
\newblock \emph{Communications of the ACM}, 38\penalty0 (11):\penalty0 39--41, 1995.

\bibitem[Nilsback \& Zisserman(2008)Nilsback and Zisserman]{nilsback2008automated}
Maria-Elena Nilsback and Andrew Zisserman.
\newblock Automated flower classification over a large number of classes.
\newblock In \emph{2008 Sixth Indian conference on computer vision, graphics \& image processing}, pp.\  722--729. IEEE, 2008.

\bibitem[Oquab et~al.(2023)Oquab, Darcet, Moutakanni, Vo, Szafraniec, Khalidov, Fernandez, Haziza, Massa, El-Nouby, et~al.]{oquab2023dinov2}
Maxime Oquab, Timoth{\'e}e Darcet, Th{\'e}o Moutakanni, Huy Vo, Marc Szafraniec, Vasil Khalidov, Pierre Fernandez, Daniel Haziza, Francisco Massa, Alaaeldin El-Nouby, et~al.
\newblock Dinov2: Learning robust visual features without supervision.
\newblock \emph{arXiv preprint arXiv:2304.07193}, 2023.

\bibitem[Radford et~al.(2021)Radford, Kim, Hallacy, Ramesh, Goh, Agarwal, Sastry, Askell, Mishkin, Clark, et~al.]{radford2021learning}
Alec Radford, Jong~Wook Kim, Chris Hallacy, Aditya Ramesh, Gabriel Goh, Sandhini Agarwal, Girish Sastry, Amanda Askell, Pamela Mishkin, Jack Clark, et~al.
\newblock Learning transferable visual models from natural language supervision.
\newblock In \emph{International conference on machine learning}, pp.\  8748--8763. PMLR, 2021.

\bibitem[Ramesh et~al.(2021)Ramesh, Pavlov, Goh, Gray, Voss, Radford, Chen, and Sutskever]{ramesh2021zero}
Aditya Ramesh, Mikhail Pavlov, Gabriel Goh, Scott Gray, Chelsea Voss, Alec Radford, Mark Chen, and Ilya Sutskever.
\newblock Zero-shot text-to-image generation.
\newblock In \emph{International conference on machine learning}, pp.\  8821--8831. Pmlr, 2021.

\bibitem[Reid et~al.(2024)Reid, Savinov, Teplyashin, Lepikhin, Lillicrap, Alayrac, Soricut, Lazaridou, Firat, Schrittwieser, et~al.]{reid2024gemini}
Machel Reid, Nikolay Savinov, Denis Teplyashin, Dmitry Lepikhin, Timothy Lillicrap, Jean-baptiste Alayrac, Radu Soricut, Angeliki Lazaridou, Orhan Firat, Julian Schrittwieser, et~al.
\newblock Gemini 1.5: Unlocking multimodal understanding across millions of tokens of context.
\newblock \emph{arXiv preprint arXiv:2403.05530}, 2024.

\bibitem[Saito et~al.(2023)Saito, Sohn, Zhang, Li, Lee, Saenko, and Pfister]{saito2023pic2word}
Kuniaki Saito, Kihyuk Sohn, Xiang Zhang, Chun-Liang Li, Chen-Yu Lee, Kate Saenko, and Tomas Pfister.
\newblock Pic2word: Mapping pictures to words for zero-shot composed image retrieval.
\newblock In \emph{Proceedings of the IEEE/CVF Conference on Computer Vision and Pattern Recognition}, pp.\  19305--19314, 2023.

\bibitem[Salehi et~al.(2024)Salehi, Farajtabar, Horton, Faghri, Pouransari, Vemulapalli, Tuzel, Farhadi, Rastegari, and Mehta]{clip-model-zoo-experts}
Mohammadreza Salehi, Mehrdad Farajtabar, Maxwell Horton, Fartash Faghri, Hadi Pouransari, Raviteja Vemulapalli, Oncel Tuzel, Ali Farhadi, Mohammad Rastegari, and Sachin Mehta.
\newblock Clip meets model zoo experts: Pseudo-supervision for visual enhancement.
\newblock In \emph{Transactions on Machine Learning Research (TMLR)}, 2024.
\newblock URL \url{https://arxiv.org/abs/2310.14108v1}.

\bibitem[Schuhmann et~al.(2022)Schuhmann, Beaumont, Vencu, Gordon, Wightman, Cherti, Coombes, Katta, Mullis, Wortsman, et~al.]{schuhmann2022laion}
Christoph Schuhmann, Romain Beaumont, Richard Vencu, Cade Gordon, Ross Wightman, Mehdi Cherti, Theo Coombes, Aarush Katta, Clayton Mullis, Mitchell Wortsman, et~al.
\newblock Laion-5b: An open large-scale dataset for training next generation image-text models.
\newblock \emph{Advances in Neural Information Processing Systems}, 35:\penalty0 25278--25294, 2022.

\bibitem[Su et~al.(2022)Su, Shi, Kasai, Wang, Hu, Ostendorf, Yih, Smith, Zettlemoyer, and Yu]{su2022one}
Hongjin Su, Weijia Shi, Jungo Kasai, Yizhong Wang, Yushi Hu, Mari Ostendorf, Wen-tau Yih, Noah~A Smith, Luke Zettlemoyer, and Tao Yu.
\newblock One embedder, any task: Instruction-finetuned text embeddings.
\newblock \emph{arXiv preprint arXiv:2212.09741}, 2022.

\bibitem[Sun et~al.(2023)Sun, Fang, Wu, Wang, and Cao]{sun2023eva}
Quan Sun, Yuxin Fang, Ledell Wu, Xinlong Wang, and Yue Cao.
\newblock Eva-clip: Improved training techniques for clip at scale.
\newblock \emph{arXiv preprint arXiv:2303.15389}, 2023.

\bibitem[Tong et~al.(2024{\natexlab{a}})Tong, Brown, Wu, Woo, Middepogu, Akula, Yang, Yang, Iyer, Pan, et~al.]{tong2024cambrian}
Shengbang Tong, Ellis Brown, Penghao Wu, Sanghyun Woo, Manoj Middepogu, Sai~Charitha Akula, Jihan Yang, Shusheng Yang, Adithya Iyer, Xichen Pan, et~al.
\newblock Cambrian-1: A fully open, vision-centric exploration of multimodal llms.
\newblock \emph{arXiv preprint arXiv:2406.16860}, 2024{\natexlab{a}}.

\bibitem[Tong et~al.(2024{\natexlab{b}})Tong, Liu, Zhai, Ma, LeCun, and Xie]{tong2024eyes}
Shengbang Tong, Zhuang Liu, Yuexiang Zhai, Yi~Ma, Yann LeCun, and Saining Xie.
\newblock Eyes wide shut? exploring the visual shortcomings of multimodal llms.
\newblock In \emph{Proceedings of the IEEE/CVF Conference on Computer Vision and Pattern Recognition}, pp.\  9568--9578, 2024{\natexlab{b}}.

\bibitem[Vani et~al.(2024)Vani, Nguyen, Lavoie, Krishna, and Courville]{vani2024sparo}
Ankit Vani, Bac Nguyen, Samuel Lavoie, Ranjay Krishna, and Aaron Courville.
\newblock Sparo: Selective attention for robust and compositional transformer encodings for vision.
\newblock \emph{arXiv preprint arXiv:2404.15721}, 2024.

\bibitem[Vaze et~al.(2023)Vaze, Carion, and Misra]{vaze2023genecis}
Sagar Vaze, Nicolas Carion, and Ishan Misra.
\newblock Genecis: A benchmark for general conditional image similarity.
\newblock In \emph{Proceedings of the IEEE/CVF Conference on Computer Vision and Pattern Recognition}, pp.\  6862--6872, 2023.

\bibitem[Vemulapalli et~al.()Vemulapalli, Pouransari, Faghri, Mehta, Farajtabar, Rastegari, and Tuzel]{vemulapalliknowledge}
Raviteja Vemulapalli, Hadi Pouransari, Fartash Faghri, Sachin Mehta, Mehrdad Farajtabar, Mohammad Rastegari, and Oncel Tuzel.
\newblock Knowledge transfer from vision foundation models for efficient training of small task-specific models.
\newblock In \emph{Forty-first International Conference on Machine Learning}.

\bibitem[Wang et~al.(2024)Wang, Vasu, Faghri, Vemulapalli, Farajtabar, Mehta, Rastegari, Tuzel, and Pouransari]{wang2024sam}
Haoxiang Wang, Pavan Kumar~Anasosalu Vasu, Fartash Faghri, Raviteja Vemulapalli, Mehrdad Farajtabar, Sachin Mehta, Mohammad Rastegari, Oncel Tuzel, and Hadi Pouransari.
\newblock Sam-clip: Merging vision foundation models towards semantic and spatial understanding.
\newblock In \emph{Proceedings of the IEEE/CVF Conference on Computer Vision and Pattern Recognition}, pp.\  3635--3647, 2024.

\bibitem[Wei et~al.(2021)Wei, Bosma, Zhao, Guu, Yu, Lester, Du, Dai, and Le]{wei2021finetuned}
Jason Wei, Maarten Bosma, Vincent~Y Zhao, Kelvin Guu, Adams~Wei Yu, Brian Lester, Nan Du, Andrew~M Dai, and Quoc~V Le.
\newblock Finetuned language models are zero-shot learners.
\newblock \emph{arXiv preprint arXiv:2109.01652}, 2021.

\bibitem[Wu et~al.(2021)Wu, Gao, Guo, Al-Halah, Rennie, Grauman, and Feris]{wu2021fashion}
Hui Wu, Yupeng Gao, Xiaoxiao Guo, Ziad Al-Halah, Steven Rennie, Kristen Grauman, and Rogerio Feris.
\newblock Fashion iq: A new dataset towards retrieving images by natural language feedback.
\newblock In \emph{Proceedings of the IEEE/CVF Conference on computer vision and pattern recognition}, pp.\  11307--11317, 2021.

\bibitem[Xu et~al.(2023)Xu, Xie, Tan, Huang, Howes, Sharma, Li, Ghosh, Zettlemoyer, and Feichtenhofer]{xu2023demystifying}
Hu~Xu, Saining Xie, Xiaoqing~Ellen Tan, Po-Yao Huang, Russell Howes, Vasu Sharma, Shang-Wen Li, Gargi Ghosh, Luke Zettlemoyer, and Christoph Feichtenhofer.
\newblock Demystifying clip data.
\newblock \emph{arXiv preprint arXiv:2309.16671}, 2023.

\bibitem[Ypsilantis et~al.(2023)Ypsilantis, Chen, Cao, Lipovsk{\`y}, Dogan-Sch{\"o}nberger, Makosa, Bluntschli, Seyedhosseini, Chum, and Araujo]{ypsilantis2023towards}
Nikolaos-Antonios Ypsilantis, Kaifeng Chen, Bingyi Cao, M{\'a}rio Lipovsk{\`y}, Pelin Dogan-Sch{\"o}nberger, Grzegorz Makosa, Boris Bluntschli, Mojtaba Seyedhosseini, Ond{\v{r}}ej Chum, and Andr{\'e} Araujo.
\newblock Towards universal image embeddings: A large-scale dataset and challenge for generic image representations.
\newblock In \emph{Proceedings of the IEEE/CVF International Conference on Computer Vision}, pp.\  11290--11301, 2023.

\bibitem[Zhang et~al.(2024)Zhang, Luan, Hu, Lee, Qiao, Chen, Su, and Chang]{zhang2024magiclens}
Kai Zhang, Yi~Luan, Hexiang Hu, Kenton Lee, Siyuan Qiao, Wenhu Chen, Yu~Su, and Ming-Wei Chang.
\newblock Magiclens: Self-supervised image retrieval with open-ended instructions.
\newblock \emph{arXiv preprint arXiv:2403.19651}, 2024.

\bibitem[Zhou et~al.(2022)Zhou, Loy, and Dai]{zhou2022extract}
Chong Zhou, Chen~Change Loy, and Bo~Dai.
\newblock Extract free dense labels from clip.
\newblock In \emph{European Conference on Computer Vision}, pp.\  696--712. Springer, 2022.

\end{thebibliography}
